%% file: main.tex
\documentclass[runningheads, orivec]{llncs}
\usepackage[T1]{fontenc}
\usepackage{graphicx}
\usepackage{multirow}
\usepackage{makecell}
\usepackage[normalem]{ulem}
\useunder{\uline}{\ul}{}
\usepackage{colortbl} 
\usepackage{array}
\usepackage[accsupp]{axessibility}
\usepackage{hyperref}
\usepackage{orcidlink}
\usepackage{accv}
\usepackage{tikz}
\usepackage[marginal]{footmisc}

\renewcommand{\thefootnote}{}
\definecolor{Gray}{gray}{0.94}

\begin{document}

\title{DFIMat: Decoupled Flexible Interactive Matting in Multi-Person Scenarios}

\author{Siyi Jiao\textsuperscript{*} \hspace{-1.5mm}$^{~\orcidlink{0009-0005-6114-1020}}$ \and Wenzheng Zeng\textsuperscript{*} \hspace{-1.5mm}$^{~\orcidlink{0000-0001-8467-5862}}$ \and Changxin Gao \hspace{-1.5mm}$^{~\orcidlink{0000-0003-2736-3920}}$ \and Nong Sang\textsuperscript{$^\dag$} \hspace{-1.5mm}$^{~\orcidlink{0000-0002-9167-1496}}$}

\titlerunning{DFIMat}
\authorrunning{Jiao et al.}

\institute{
  School of AIA, Huazhong University of Science and Technology\\
  \email{\{m202173030,m202173066\}@alumni.hust.edu.cn},
  \email{\{cgao,nsang\}@hust.edu.cn}
}

\maketitle    

\newcommand\nnfootnote[1]{%
  \begin{NoHyper}
  \renewcommand\thefootnote{}\footnote{#1}%
  \addtocounter{footnote}{-1}%
  \end{NoHyper}
}

\nnfootnote{\textsuperscript{*} Equal contribution.\\ \dag \ Corresponding author.}

\input{0_abstract} 
\input{1_intro}
\input{2_related}

\input{3_method}

\input{4_dataset}

\input{5_experiments}

\input{6_conclusion}

\input{main.bbl}
\end{document}

%% file: 0_abstract.tex
\begin{abstract}
  Interactive portrait matting refers to extracting the soft portrait from a given image that best meets the user's intent through their inputs.
Existing methods often underperform in complex scenarios, mainly due to three factors. (1) Most works apply a tightly coupled network that directly predicts matting results, lacking interpretability and resulting in inadequate modeling. (2) Existing works are limited to a single type of user input, which is ineffective for intention understanding and also inefficient for user operation. (3) The multi-round characteristics have been under-explored, which is crucial for user interaction. To alleviate these limitations, we propose DFIMat, a decoupled framework that enables flexible interactive matting. Specifically, we first decouple the task into 2 sub-ones: localizing target instances by understanding scene semantics and the flexible user inputs, and conducting refinement for instance-level matting. We observe a clear performance gain from decoupling, as it makes sub-tasks easier to learn, and the flexible multi-type input further enhances both effectiveness and efficiency. DFIMat also considers the multi-round interaction property, where a contrastive reasoning module is designed to enhance cross-round refinement.
Another limitation for multi-person matting task is the lack of training data. We address this by introducing a new synthetic data generation pipeline that can generate much more realistic samples than previous arts. A new large-scale dataset SMPMat is subsequently established. Experiments verify the significant superiority of DFIMat. With it, we also investigate the roles of different input types, providing valuable principles for users. Our code and dataset can be found at \href{https://github.com/JiaoSiyi/DFIMat}{\textcolor[RGB]{255,69,130}{https://github.com/JiaoSiyi/DFIMat}}.

\end{abstract}

%% file: 1_intro.tex
\section{Introduction}
\label{sec:introduct}


\begin{figure}[t]
  \begin{minipage}{0.63\textwidth}
    \captionof{table}{Comparison of different methods. DFIMat supports (1) multi-type user inputs, (2) any combination of different types of input at each time, and (3) multi-round iteration.}
    \resizebox{\linewidth}{!}{
    \centering
    \renewcommand{\arraystretch}{1.12}
   \begin{tabular}{c|c|cccc|c|c}
\hline
\multicolumn{1}{c|}{\multirow{2}{*}{Design}} & \multirow{2}{*}{Approach} & \multicolumn{4}{c|}{Supported input type} & \multirow{2}{*}{\begin{tabular}[c]{@{}c@{}}Mixed type\\ of input?\end{tabular}} & \multirow{2}{*}{\begin{tabular}[c]{@{}c@{}}Multi-\\ round?\end{tabular}} \\ \cline{3-6}
\multicolumn{1}{c|}{}                        &                           & click    & scribble    & box    & text    &                                    &                               \\ \hline
\multirow{4}{*}{Coupled}                     & ~\cite{DIIM,click,ugd}            & \checkmark      &             &        &         &                                    &                               \\
                                             & ~\cite{active,smart}              &          & \checkmark         &        &         &                                    & \checkmark                           \\
                                             & FGI~\cite{flex}                       & \checkmark      & \checkmark         &        &         &                                    &                               \\
                                             & RIM~\cite{rim}                      &          &             &        & \checkmark     &                                    &                               \\ \hline
\multirow{2}{*}{Decoupled}                   & MatAny~\cite{matany}                    & \checkmark      & \checkmark         & \checkmark    & \checkmark     &                                    &                               \\
                                             & DFIMat (ours)                     & \checkmark      & \checkmark         & \checkmark    & \checkmark     & \checkmark                                & \checkmark                           \\ \hline
\end{tabular}}
\label{tab:introcompare}
  \end{minipage}
  \begin{minipage}{0.34\textwidth}
    \centering
    \includegraphics[width=\linewidth]{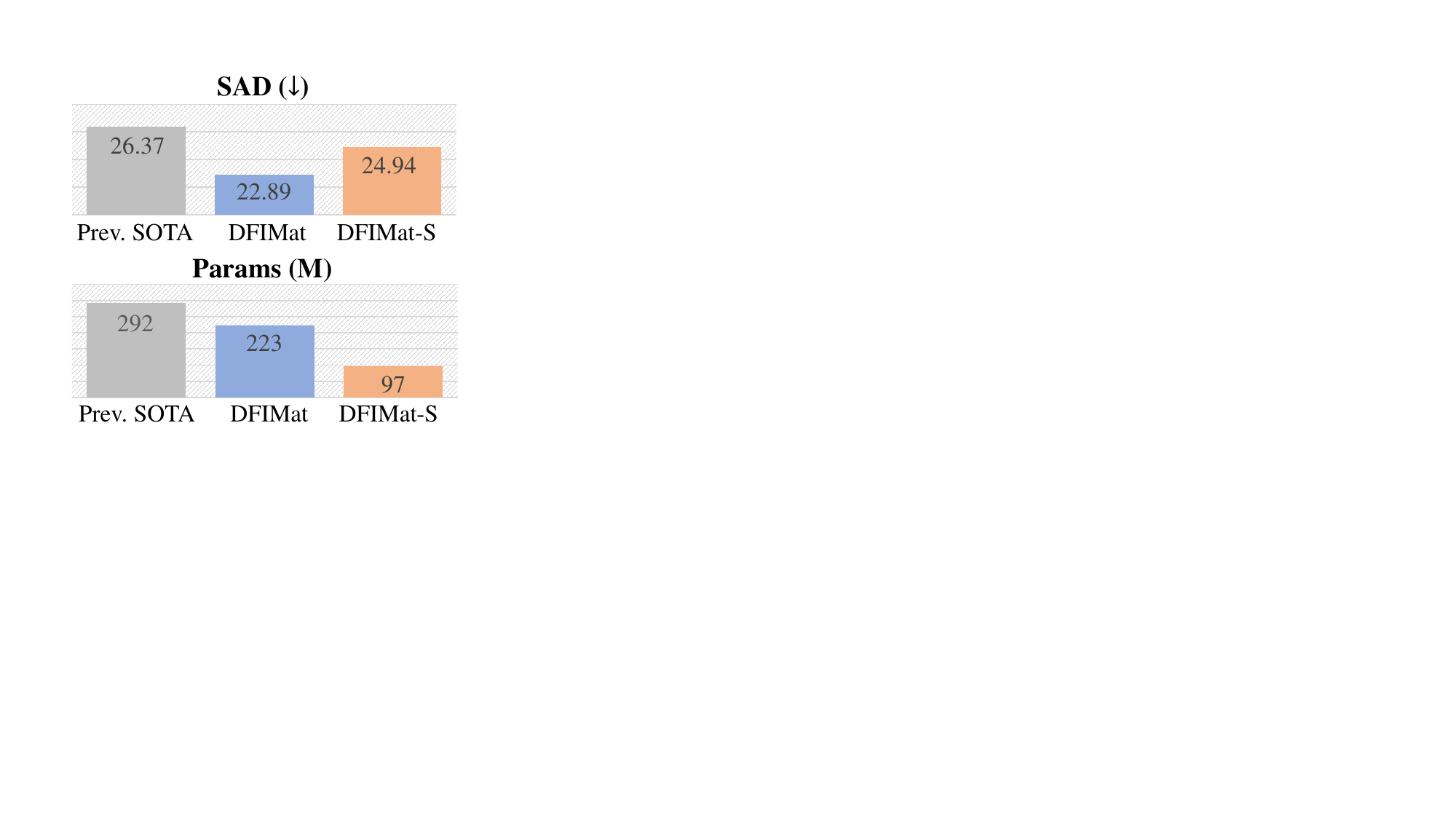}
    \captionof{figure}{Performance and complexity comparison.}
   \label{fig:introcompare}
  \end{minipage}
\end{figure}


Interactive portrait matting (IPM) is a crucial computer vision task that aims to extract the fine-grained alpha matte of the specific foreground instance that can best match the users' intention given from their interactive inputs (e.g., clicks, scribbles, texts). The significance of IPM lies in its wide downstream application values, such as image editing, advertisement production, and video conferencing.

Previous studies \cite{smart,click,active,flex,DIIM,ugd,rim}, have demonstrated successful performance validation in relatively idealized scenarios (e.g., clear background and single instance without occlusions). However, in real-world scenarios, images often present highly complex backgrounds, with multiple instances and even severe instance occlusions~\cite{inst}. As a result, the existing approach have exhibited poor performance on images that are more representative of real-world scenarios~\cite{inst}.

We argue that there are three crucial reasons that potentially cause the failure of the aforementioned methods in challenging real-world scenarios. Firstly, they generally use a coupled network to directly predict matting results. From a top-down perspective, the task actually consists of multiple steps: we need to localize the targeted instance conditioning on the understanding of both user intention and scene semantics, and then conduct fine-grained matting on the corresponding instance. The coupled design lacks interpretability and makes the network difficult to learn each sub-tasks well especially in complex scenarios.

The second issue is that existing works only consider a single type of user input for the model to understand the user intention, which is inefficient for instance matting task that need both global scene understanding for instance localization and local awareness for fine-grained matting at boundary. Specifically, the bounding box input is a suitable way to quickly localize target instance, but lack of ability to focus on fine-grained boundary details. While click and scribble inputs are good at distinguishing local fine-grained details but inefficient for instance localization. One optimal way is first using a bounding box to localize instance and then applying click and scribble to refine local boundary. Such a multiple-type input will also make the user operation more flexible. Therefore, enabling a universal matting interface that is capable of accommodating various types of human prompts is more preferred.

Thirdly, for practical usage, multiple rounds of interaction are typically necessary to refine the matting result until it is satisfactory. However, most works only support feeding user inputs in one step, ignoring the cross-round information that can be useful clues for refinement.

Based on the aforementioned analysis, we propose DFIMat, a decoupled framework that enables flexible interactive matting. Particularly, we propose to decouple the IPM task into 2 sub-ones: localizing target instances by understanding scene semantics and the flexible user inputs, and conducting refinement for instance-level matting. Following this rule, we subsequently design an interactive semantic capture network (ISCN) and a matting refinement network (MRN) to address these two tasks respectively. 

Within ISCN, we propose to enable multi-modal user inputs such as clicks, scribbles, boxes, texts, or any combination of them, resulting in a more concise, flexible, and efficient interaction. This is achieved by encoding various inputs into a unified visual-semantic space, and building strong interactions in the decoder to understand the user intents and predict the target instance mask for instance localization. To meet the practical needs, DFIMat also consider the multi-round interaction property, where we design a contrastive reasoning module to evaluate the consistency between the model prediction and user intention while also explicitly identifying and reasoning the conflict areas during each round's interaction, providing valuable auxiliary guidance for cross-around refinement. 

For MRN, we build a dual-branch network to effectively capture fine-grained local details with global instance-level consideration. As summarized in~\cref{tab:introcompare}, our DFIMat distinguishes existing works in: (1) supporting multi-types of user inputs; (2) allowing any combination of different input types (including a single input) at each time; (3) with multi-round iteration ability. Those properties make it more user-friendly and with better effectiveness as verified by experiments.

Data is another important point for method training and evaluation. The volume of real-image datasets for multi-person matting remains relatively small due to the cost of data collection and annotation. In order to obtain a large amount of matting data that contains multi-instance scenes, previous methods~\cite{inst,rim} adopt a simple synthesis strategy to iteratively add portrait foregrounds to no-portrait backgrounds. Due to the randomness of the adding positions and the lack of instance-scene prior consideration, there is often a large gap between the synthetic images and natural images, it is more preferred to utilize more realistic and complex images for training and evaluation. To fill this gap, we further design a new synthetic data generation pipeline that can generate much more diverse and realistic samples, and build a new large-scale dataset SPMMat, which consists of 40,000 realistic multi-instance images with high-quality matte GT.

Our extensive experiments verify the superiority of DFIMat over representative methods. Notably, DFIMat outperforms previous SOTA by 3.48 SAD on the challenging SMPMat dataset with higher efficiency. We also provide a more lightweight version, DFIMat-S,  with only 33\% of the parameters of SOTA methods, while still achieving higher matting accuracy, as shown in~\cref{fig:introcompare}. By utilizing DFIMat, we also investigate the roles of different input types, providing valuable principles for users on more effective interaction. Our main contributions are:

\begin{itemize}
    \item We propose a decoupled network for IPM task, which decomposes the task based on a top-down perspective, resulting in a clear performance gain.
    \item We propose to enable flexible and multi-type user input for interactive matting, making it more effective, efficient, and user-friendly. This is achieved by encoding different inputs into a unified visual-semantic space. 
    \item Concerning the multi-round feature of interaction, we design a contrastive reasoning module to enhance cross-round refinement.
    \item We propose a new synthetic data generation pipeline that can generate diverse and high-quality image-matte-text pairs in multi-person scenarios. A large-scale dataset is further introduced to facilitate relevant research.
    \item We investigate the roles of different input types and provide valuable principles for users on more effective interaction.
\end{itemize}

%% file: 2_related.tex
\section{Related Work}
\label{sec:Related}

\paragraph{Interactive Image matting.} Existing methods~\cite{click,smart,active,flex,DIIM,ugd,rim} adopt user inputs to identify the foreground and background region, which can usually obtain much better matting results than the automatic ones~\cite{dim,late,mfor,learning,ma2023rethinking,18semantic,deep,boo,mod,bri,p3m,attention,robust,indexnet,tri,gca,context,tangle,gan}. 
Most of the existing interactive matting methods~\cite{active,DIIM,click,smart,flex,ugd,rim} adopt an encoder-decoder-like architecture that takes image and user input as input, and directly predicts matting results. Such a coupled network design makes the model difficult to adapt well in complex real-world scenarios, such as multi-person scenes with severe occlusions. The reason is that the coupled design lacks interpretability and thus increasing the learning difficulty. 

Another limitation is that existing works generally only consider a single type of user input (e.g., click, scribble, box, or text), which is ineffective and not user-friendly, as different types of inputs can play different roles and can complement each other. Although a very recent work (i.e., MatAny~\cite{matany}) expands the input type by using a segmentation foundation model (SAM~\cite{sam}) to receive different types of input, it still can not support mixed types of input at one time, which failed to exploit complement information from different types of input during user interactions. Moreover, most of the existing works do not consider the multi-round interaction property that is necessary for practical usage and thus ignore the cross-round information that can be useful clues for refinement. 

Here we proposed a decoupled network DFIMat that is of better interpretability and performance. It also enables truly flexible inputs by encoding different types of inputs into a unified visual-semantic space, resulting in a more effective and user-friendly matting experience. We also consider the multi-round interaction characteristic and design a contrastive reasoning module to enhance cross-round refinement. A summary of different methods can be seen in~\cref{tab:introcompare}.

\paragraph{Matting datasets.} Numerous matting datasets~\cite{dim,p3m,attention} have been introduced to propel advancements in the field of image matting. Typical matting datasets contain high-resolution images belonging to some specific object categories that have lots of details like hair, accessories, fur, and net, as well as transparent objects. Besides, some other matting datasets focus on a specific category of object, e.g., humans~\cite{tri,p3m} and animals~\cite{aim}. To generate images containing multiple foreground objects, a typical solution in previous matting methods~\cite{inst,rim} is to iteratively composite the foreground onto the background image sequentially. Although augmentation strategies have been proposed to reduce the domain gap between the real-world images and the composite ones,there is still an urgent need to generate synthetic images that are closer to the real world. 

In our work, we build a multi-person matting dataset and ensure its diversity and high quality by designing a new synthetic data generation pipeline.


%% file: 3_method.tex
\section{Method}

\begin{figure*}[t]
    \centering
    \includegraphics[width=1\linewidth]{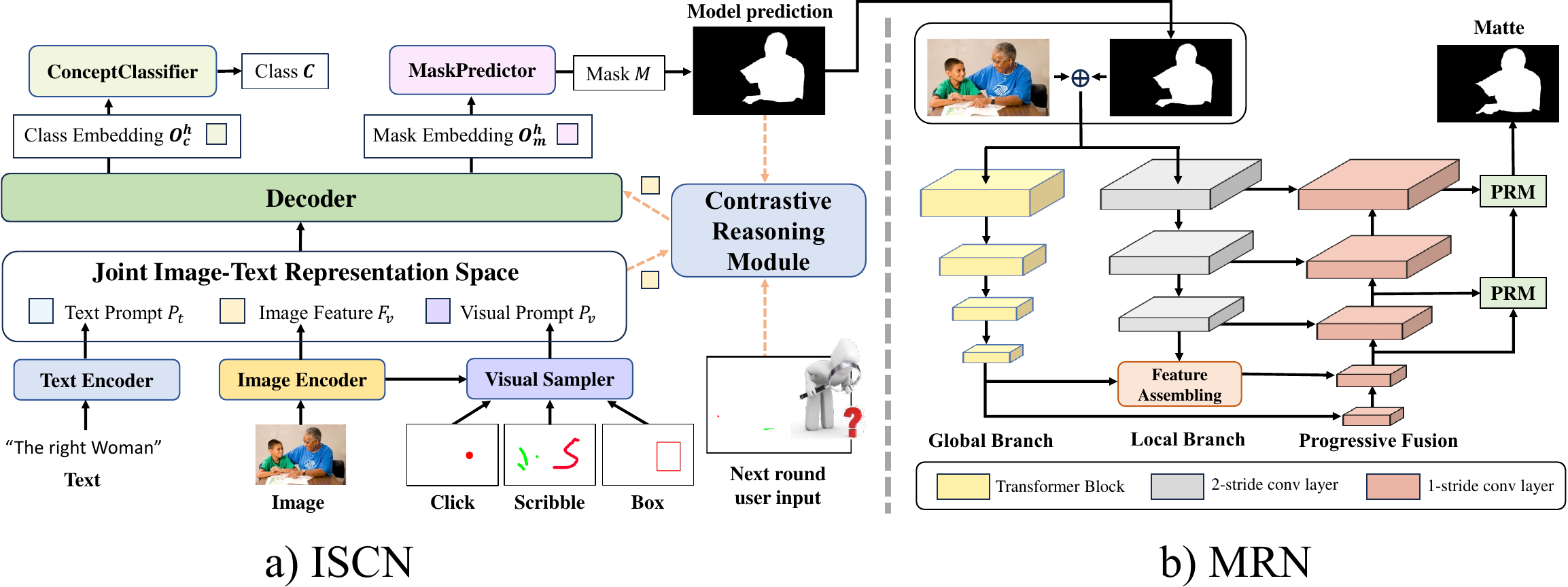}

    \caption{The overall framework of DFIMat, which consists of two components: (a) Interactive semantic capture network (ISCN), and (b) Matting refinement network (MRN).}
    \label{fig:frame1}
\end{figure*}

\subsection{Overview} \label{subsec:overview}

By reflecting on and summarizing the shortcomings of the existing works, we propose DFIMat, a novel decoupled framework for flexible interactive matting. It consists of two independent components: the interactive semantic capture network (ISCN) and the matting refinement network (MRN), as illustrated in~\cref{fig:frame1}. The ISCN is responsible for understanding the user intention from their various inputs, and localizing the interested instance in the image (\cref{subsec:SCE}). The MRN takes the prediction result of ISCN (i.e., a coarse mask of the interested instance) as well as the original image, and performs refinement to produce the final alpha matte for the corresponding instance-level matting (\cref{subsec:MRN}).


 \subsection{Interactive Semantic Capture Network (ISCN)} \label{subsec:SCE}
Here we introduce the proposed ISCN, a model that understands scene semantics and multiple/flexible user inputs to localize the target instance, in an interactive manner. The ISCN design takes inspiration from the recent success of multi-modal learning methods ~\cite{mdetr,maple,seem} and encodes the different types of user inputs and image into a unified visual-semantic space. Then, strong interactions among them are introduced to better understand the user intentions and then predict the target instance mask for instance localization. Leveraging the characteristic of multi-round interaction, we further design a simple but effective contrastive reasoning module to evaluate the consistency between the model prediction and user intention while also explicitly identifying and reasoning the conflict areas during each round's interaction, providing valuable auxiliary guidance for the model to further refine its outputs. The overall architecture is shown in~\cref{fig:frame1}, which will be introduced in detail.

\noindent\textbf{Unified visual-semantic space.} 
Given an input image $I \in \mathbb{R}^{H \times W \times 3}$, we first extract image feature $F_v$ by an image encoder. Visual inputs $s$($s \in\{clicks, scrib$ $-bles, boxes\}$) are converted to visual prompts $P_v$ through a visual sampler:
\begin{equation}
P_v = \mathbf{VisualSampler}(s,F_v).
\label{eq:sample}
\end{equation}
The visual sampler performs feature point sampling on the corresponding locations in image features based on user inputs. Textual inputs are fed into a text encoder for text prompts $P_t$.
Then, a decoder is used to build strong interactions between the image feature $F_v$ and user prompts ${P_v, P_t}$ to understand both scene semantics and user intentions, and finally predict the target instance mask for instance localization. This can be formulated as follows:
\begin{equation}
\left\langle{M}, {C}\right\rangle  =\mathbf{Decoder}\left(\left\langle{P}_t, {P}_v\right\rangle \mid {F}_v\right), 
\label{eq:decoder}
\end{equation}
where $M$ is the predicted instance mask and $C$ is its class (is human or not). More specifically, we first initialize three types of learnable query: object queries $Q_o$, text queries $Q_t$ and visual queries $Q_v$. Each decoder stage contains cross-attention operations on images and learnable queries, and a prompt self-attention block to perform the interaction between queries and prompts:
\begin{equation}
\begin{aligned}
&Q_x = \mathbf{CrossAttention}(Q=F_v,K=V=Q_x),\quad x \in \{o,v,t\},\\
&Q_o,Q_t,Q_v = \mathbf{SelfAttention}(Q_o,Q_t,Q_v,P_v,P_t).
\end{aligned}
\end{equation}
The output of last decoder stage and image features are passed to FFN to obtain the mask embeddings ${O}_h^m$ and class embeddings ${O}_h^c$:
\begin{equation}
{O}_h^m,{O}_h^c = \mathbf{FFN}(F_v,Q_o,Q_t,Q_v).
\end{equation}
Finally, ISCN predicts the masks $M$ and the classes $C$  based on ${O}_h^m$ and ${O}_h^c$:
\begin{equation}
\begin{aligned}
\mathbf{M} &=\mathbf{MaskPredictor}\left(\mathbf{O}_h^m\right),\\
\mathbf{C} &=\mathbf{ConceptClassifier}\left(\mathbf{O}_h^c\right).
\end{aligned}
\label{eq:mc}
\end{equation}
Here MaskPredictor (and ConceptClassifier) are task-specific heads and we follow the design of X-decoder~\cite{zou2023generalized} for its simplicity.


\noindent\textbf{Contrastive reasoning module.} We propose a contrastive reasoning module (CRM) to evaluate the consistency between the model prediction and user intention while also explicitly identifying and reasoning the conflict areas during each round's interaction, which can serve as a useful clue for cross-round refinement. Specifically, we maintain a mask $M_{ref} \in \mathbb{R}^{H \times W}$, where each pixel has one of the following three values based on the difference between new user input and the previous model prediction: (1) $D_{r} = 0$ indicates no conflict; (2) $D_{fg} = 1$ means previous prediction classified the area as background but new user input suggest it is foreground, and (3) $D_{gf} =2$ means conversely with $D_{fg}$. $M_{ref}$ is initialized with all pixels set to $D_{r}$ and recalculated upon receiving new user input. A convolution layer is utilized to transform $M_{ref}$ into an embedding $E_c \in \mathbb{R}^{H \times W \times C_1}$ , with each pixel value corresponding to a learnable $C1$-dimensional vector. Then, we resize and combine $E_c$ with the image feature $F_v$ to get the conflict-involved feature ${F_v} = resize(E_c)+F_v$ for cross-round refinement. In addition, the previous prediction result is also sent to the decoder as a mask to participate in the calculation of masked multi-head attention in prompt self-attention block.

\noindent\textbf{Loss.}
We train ISCN with standard segmentation loss:
\begin{equation}
\mathcal{L}_{\text {ISCN}} = \mathcal{L}_{c\_ce}(C,\hat{C}) + \mathcal{L}_{m\_bce}(M,\hat{M}) + \mathcal{L}_{m\_dice}(M,\hat{M}),
\label{eq:loss1}
\end{equation}
where $\hat{C}, \hat{M}$ is GT category and mask respectively. $\mathcal{L}_{c\_ce}$, $\mathcal{L}_{m\_bce}$, and $\mathcal{L}_{m\_dice}$ denote cross-entropy, binary cross-entropy, and dice loss, with weights 0.1:1:1.

\subsection{Matting Refinement Network (MRN)} \label{subsec:MRN}
MRN aims to refine the mask prediction from ISCN to obtain accurate alpha matte predictions. As ISCN has already given a relatively good instance mask as a good beginning, the task difficulty for MRN is largely reduced.  
Here, we design a simple dual-branch network as our MRN, to capture fine-grained local details while simultaneously considering global instance-level semantics, as shown in~\cref{fig:frame1}. Our insight is that images for multi-instance matting tasks often contain complex human interaction and background, so a global encoder is needed to better capture the overall structure and background information, while a local encoder can focus more on details. Then, a subsequent progressive feature fusion should be built to fuse the features and decode them to final matte.
Taking those things in mind, we design a simple network containing a global encoder, a local encoder, and a progressive feature fusion module, as shown in ~\cref{fig:frame1}. Specifically, we choose a CNN as the local encoder, as it can effectively exploit local features, and we choose a transformer-based encoder as our global encoder, as the self-attention operation can build strong non-local interactions in the images to form a better global instance-level understanding. For progressive fusion, we start from the latent feature from the global branch, as it contains rich global instance-level representation. We then fuse it with the feature from local branch in a progressive manner (from low to high resolution), we also utilize PRM~\cite{MG} to further refine the result.

\noindent\textbf{Loss.} Since most pixels in the coarse mask are already predicted correctly, only a few ``hard” pixels need significant refinement. To make the network pay more attention to those ``hard” pixels, we adopt a simplified hard-sample mining objective function $\mathcal{L}_{\text {MRN}}$ as follows:
\begin{equation}
\setlength{\abovedisplayskip}{3pt}
\setlength{\belowdisplayskip}{3pt}
\mathcal{L}_{\text {MRN}}=\frac{1}{|C|} \sum_{i \in C}\left|\alpha_p^i-\alpha_g^i\right|+\lambda \frac{1}{|H|} \sum_{j \in H}\left|\alpha_p^j-\alpha_g^j\right|,
\label{eq:loss2}
\end{equation}
where $C$ represents the whole pixel-set and $H$ denotes ``hard” pixel-set whose error to corresponding ground truth ranks in the top 30\% of the matte. $\lambda$ denotes the weight that emphasizes the hard samples and is set as 1 by default.

%% file: 4_dataset.tex
\section{The SMPMat Dataset}

We propose a synthetic multi-person matting dataset called SMPMat to facilitate the research of instance matting task. To our knowledge, the existing multi-instance dataset from natural images~\cite{inst} suffers from low data scale as well as diversity. Specifically, HIM2K~\cite{inst} is the mainly used dataset that focuses on instance-level matting under multi-person scenarios. As can be observed in~\cref{tab:dataset_com}, the HIM2K dataset only contains a limited scale of data that is collected from natural scenes  (i.e., only contains 320 natural images with 930 instances in total), making it only serve as a validation set. 

\begin{figure*}[t]
\begin{subfigure}{0.48\linewidth}
      \includegraphics[width=\linewidth]{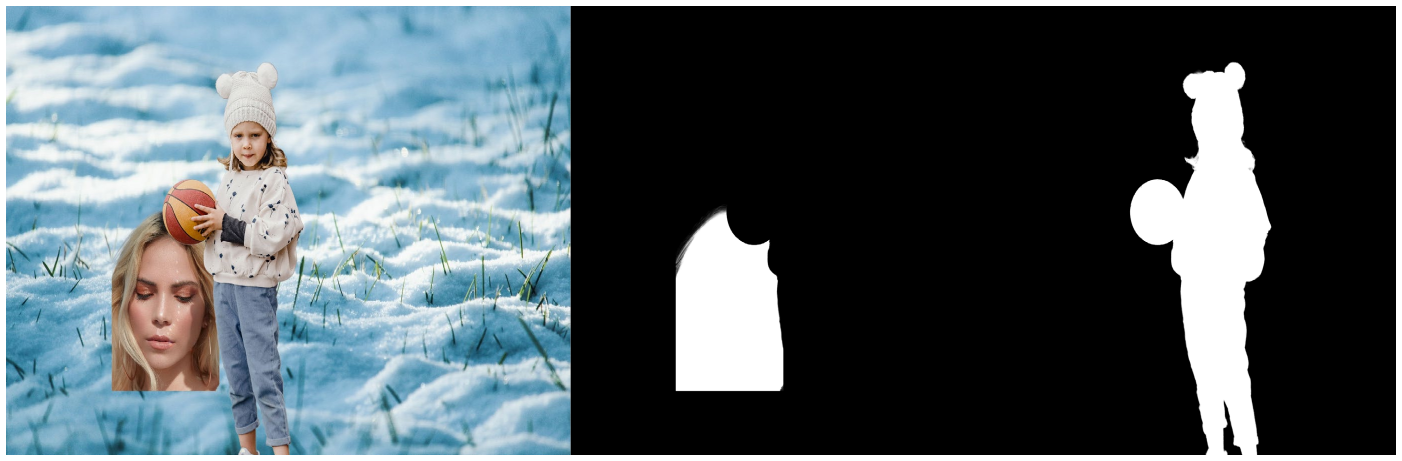}
      \caption{Existing synthetic method~\cite{inst}.}
    \end{subfigure}
\begin{subfigure}{0.462\linewidth}
      \includegraphics[width=\linewidth]{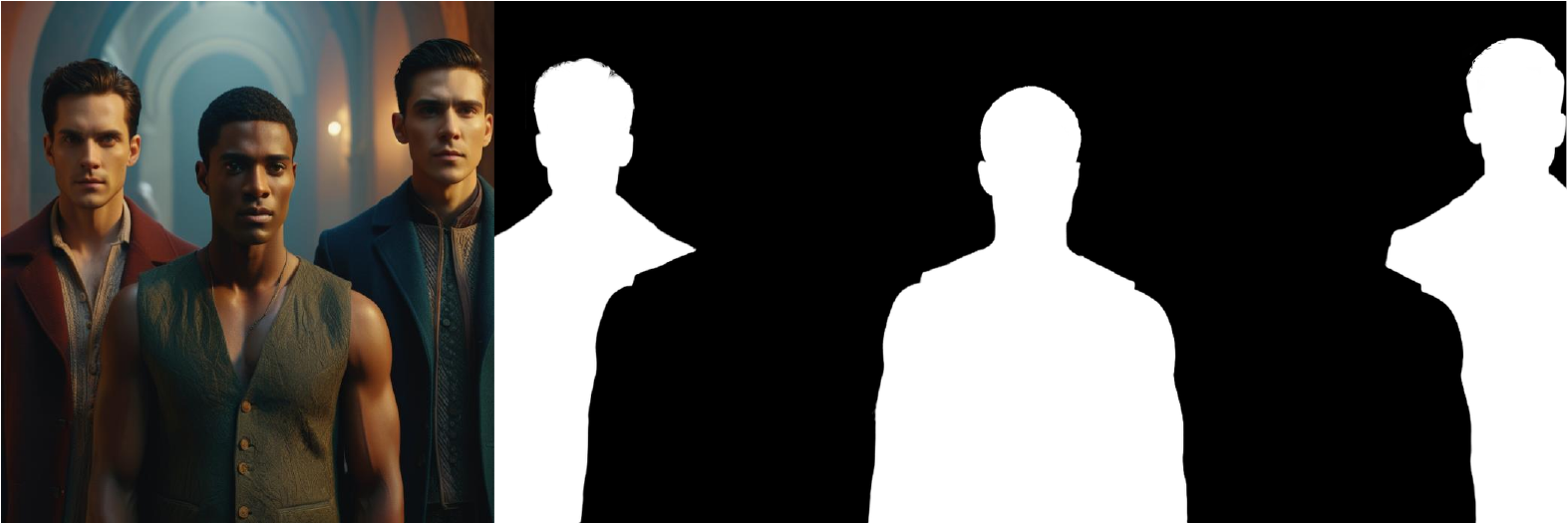}
      \caption{Our method}
    \end{subfigure}   
\caption{Visual comparison of synthetic datasets.}
\label{fig:dataset_com}
\end{figure*}

In order to obtain a large amount of matting data that contains multi-instance scenes, previous methods~\cite{inst,rim} adopt a simple synthesis strategy to iteratively add portrait foregrounds to no-portrait background. Due to the randomness of the adding positions and the lack of instance-scene prior consideration, there is often a large gap between the synthetic images and natural images, as shown in~\cref{fig:dataset_com}. To fill this gap, we design a new synthetic data generation pipeline that can generate much more diverse and realistic samples, and build a new large-scale dataset SMPMat, which consists of 40,000 realistic multi-instance images with high-quality matte GT.

\begin{table}[t]
\centering
\caption{Comparison with previous multi-instance matting dataset and ours.}
\begin{tabular}{c|c|c|cc}
\Xhline{1pt}
\multirow{2}{*}{Datasets} & \multirow{2}{*}{Image Number} & \multirow{2}{*}{Instance Number} & \multicolumn{2}{c}{Annotation}                     \\ \cline{4-5} 
                          &                               &                                  & \multicolumn{1}{c|}{Matte}      & Text description \\ \hline
HIM2K(nature)             & 320                           & 830                              & \multicolumn{1}{c|}{\checkmark} &                  \\
HIM2K(synthetic)          & 1,680                         & 5,884                            & \multicolumn{1}{c|}{\checkmark} &                  \\
SMPMat                    & 40,000                        & 142,357                          & \multicolumn{1}{c|}{\checkmark} & \checkmark       \\ \Xhline{1pt}
\end{tabular}
\label{tab:dataset_com}
\end{table}

\noindent\textbf{The synthetic data generation pipeline.}
Inspired by work~\cite{data1,data2,dm} using diffusion models for image synthesis, we build a new synthetic data generation pipeline that can synthesize an infinite amount of realistic and diverse images with high-quality matte ground truth. Our insight is that the latent feature within the well-trained diffusion process (e.g., Stable Diffusion~\cite{data1}) contains rich semantic contexts of the corresponding generated image, so ideally, the matte ground truth of the generated image can be well interpreted from such latent feature. Thus, we build an interpreter to derive the matte ground truth from the latent feature of the diffusion process. Based on the good performance of the recent text-to-image diffusion models, all we need to do is just train the interpreter for this interpretation task, using a small amount of annotated real data. After that, we can synthesize an infinite amount of images with high diversity based on the well-trained text-to-image diffusion models (e.g., Stable Diffusion~\cite{data1}), and simultaneously obtain their matte ground truth by our trained interpreter.

\begin{figure}[t]
  \begin{minipage}{1\textwidth}
    \centering
    \begin{subfigure}{\linewidth}
      \centering
      \includegraphics[width=1\linewidth]{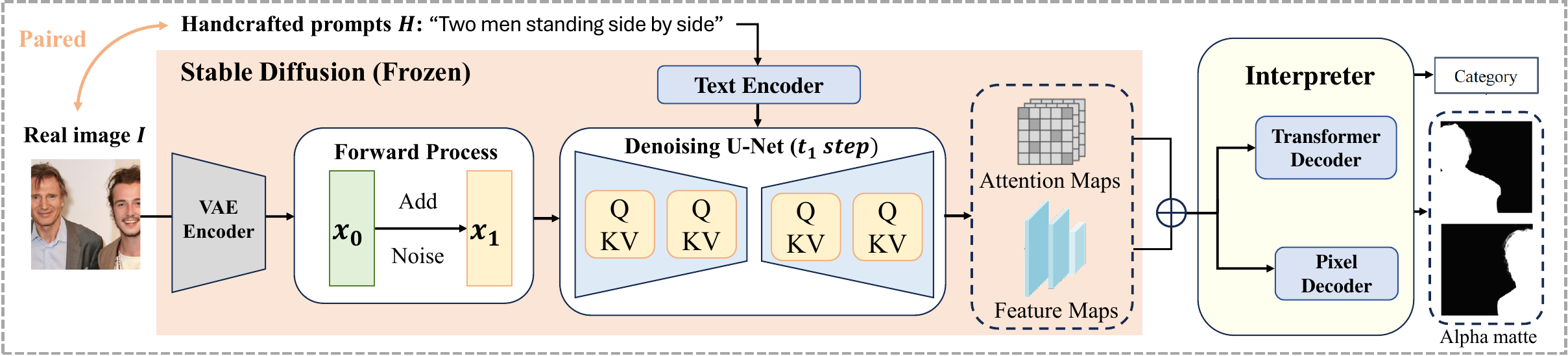}
      \caption{Training phase with few (400 images) manually labeled natural images.}
    \end{subfigure}
    \begin{subfigure}{\linewidth}
      \centering
      \includegraphics[width=1\linewidth]{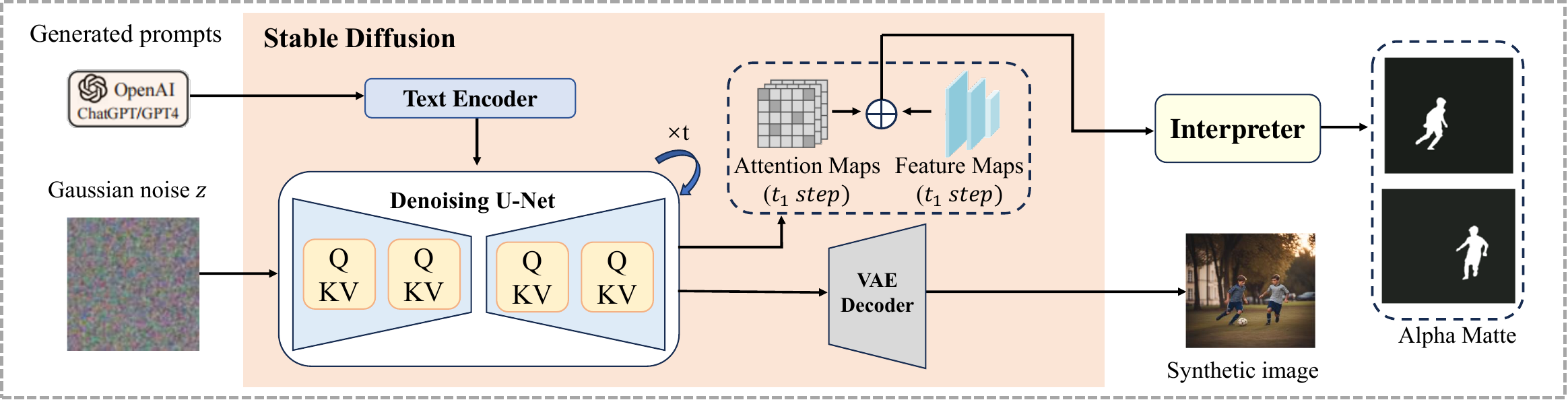}
      \caption{Inference phase generating image and corresponding perception annotation.}
    \end{subfigure}
    \caption{The synthetic data generation pipeline.}
    \label{fig:data_gen}
  \end{minipage}
\end{figure}

Specifically, our method is illustrated in~\cref{fig:data_gen}. During training, given a real image $I$ and the paired text description $H$, we feed them into a pre-trained text-to-image diffusion model (i.e., Stable Diffusion~\cite{data1} in our implementation) and acquire the multi-scale latent feature map $\mathcal{F}$ as well as the text-visual cross-attention map $\mathcal{M}$ in it (i.e., the denoising U-Net). 
We concat those intermediate representations $\hat{\mathcal{F}} = Concat(\left[\mathcal{F}, \mathcal{M}\right])$ and send $\hat{\mathcal{F}}$ to our interpreter and interpret them into GT matte. For the detailed architecture of the interpreter, any decoder for dense prediction task can be used, and here we adopt Mask2Former~\cite{mask2former}, which contains a transformer decoder and a pixel decoder. Given $\hat{\mathcal{F}}$, and $N$ learnable queues $\{Q_0, Q_1...Q_T \}$ as input, it outputs $N$ foreground alpha matte $A$ and their corresponding categories $L$ (is a human instance or not). We train the interpreter with binary cross-entropy loss and alpha loss:
\begin{equation}
\mathcal{L}_{\text {P-decoder}}=\lambda \mathcal{L}_{c\_bce}(\hat{L},L)+\mathcal{L}_{a\_alpha}(\hat{A},A),
\end{equation}
where $\lambda=0.1$, $\hat{L}, \hat{A}$ means the ground truth category and alpha matte respectively. To enable the interpreter training, we collected 400 real images and labeled them with matte ground truth as well as a text description. Once the training finishes, we use the Stable Diffusion to generate realistic images, and use the trained interpreter to obtain the matte ground truth at the same time. To let the Stable Diffusion generate diverse images, we design a prompt to instruct GPT-4~\cite{gpt} to generate an infinite amount of diverse and semantic-rich text descriptions (see supplementary for details), and send them to the Stable Diffusion for text-to-image generation. We give a visual comparison in~\cref{fig:dataset_com}. It can be observed that the image synthesized by existing data synthetic algorithm~\cite{inst} often lacks of realistic instance lay-out with scene-instance prior consideration, while our new data synthetic pipeline enables a much more realistic data generation. Please also refer to~\cref{sec: quantitative_data_gen} for quantitative evaluation.

\noindent \textbf{The SMPMat dataset.}
We followed the above generation pipeline to generate a large-scale multi-person matting dataset SMPMat, in which we carefully select 40,000 high-quality multi-person scene images from our generated samples to form the dataset. We generated diverse text descriptions for the people in each image through GPT-4~\cite{gpt}, as the additional text annotations to broaden the usage of the proposed dataset. 
Compared with existing multi-person matting datasets, the SMPMat dataset shows superiority in both data diversity (see~\cref{tab:dataset_com}) and image quality (see~\cref{fig:dataset_com}).

%% file: 5_experiments.tex
\section{Experiments}

\subsection{Implementation Details}

We first train the semantic capture network alone without the matting refinement network. After the semantic network converges, we freeze it and then train the matting refinement network. For all the network training, Adam optimizer~\cite{adam} is used and the base learning rate is set to $5 \times 10^{-4}$ with the cosine learning rate scheduler. The matting network is trained for 150 epochs, while the uncertainty estimation decoder and the refinement network are trained for 75 epochs. 


\subsection{Dataset and Evaluation Protocol} \label{sec: protocol}
We compare our method with existing interactive matting methods~\cite{flex, click, ugd, smart, DIIM, active, rim}
on the SMPMat dataset and the HIM2K (natural) dataset~\cite{inst}.
All the methods are trained on the training set of SMPMat, and evaluated on both the validation set of SMPMat and HIM2K (natural). All the used metrics (the smaller, the better) follow previous works.
We train different methods under their supported input type.
For a fair comparison, we first train our DFIMat under the same protocol as the existing method (only one type of input) to make the comparison under each supported type. Then, we also train our DFIMat using mixed types of user input to show its full performance. 
We design some rules to imitate human behavior and simulate the user input during training and testing. Rules are as follows.

\noindent\textbf{Click \& scribble input.} Since they usually conduct in a multi-round interaction fashion, we set a 5-round interaction loop, for both training and testing. Each round adds 3 clicks/2 scribbles. The first round of input is randomly generated by GT (but the same for different methods in the same input image for a fair comparison). Subsequent inputs are generated in the most significant area calculated from previous prediction results and GT (see the supplementary material for the definition of the most significant area). For methods that do not consider multi-round interactions, we aggregate the inputs from rounds 1 to t and feed them together into the model as the input for round t.

\noindent\textbf{Box \& text input.} A single round of interaction is built for both training and testing, where the input is obtained directly from GT.

\noindent\textbf{Mixed input.} For both training and testing, the interaction round is set to 5. We set the first round of input to a combination of text and any kind of visual input (click/scribble/box). The input types in $2\sim3$ rounds are randomly selected from click and scribble, and clicks/scribbles are added to the most significant areas based on previous prediction results (same rule as the aforementioned single-type click or scribble input).

\subsection{Comparison with the state-of-the-art methods}
\begin{figure}[t]
    \includegraphics[width=\linewidth]{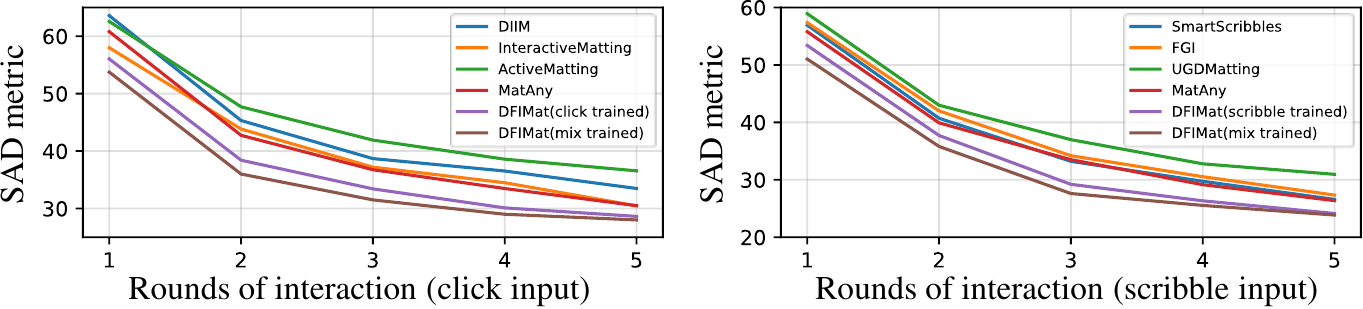}
    \captionsetup{font=footnotesize}
    \caption{Performance comparison under different rounds of interaction on SMPMat.}
     \label{fig:click&scribble}
\end{figure}

\noindent\textbf{Single-type user input.} 
We conducted comparisons of various models on the SMPMat validation set and the HIM2k natural subset, with results listed in \cref{tab:Comparison}. Experimental outcomes demonstrate that under single-type input settings, our approach consistently outperforms all state-of-the-art methods. To investigate how different models perform during multiple rounds of interaction, we present the SAD variation curves of various methods during the interactive process in \cref{fig:click&scribble}. Notably, Our DFIMat also achieves more accurate prediction output with fewer interactions needed. 

\begin{table}[t]
\caption{Quantitative comparison on SMPMat validation set and HIM2K natural set. The MSE metrics are scaled by $10^2$. }
\resizebox{\linewidth}{!}{
\begin{tabular}{c|c|cccc|cccc}
\hline
                                                                                  &                                                  & \multicolumn{4}{c|}{SMPMat Validation}                                                                                       & \multicolumn{4}{c}{HIM2K Natural}                                                                                            \\
\multirow{-2}{*}{\begin{tabular}[c]{@{}c@{}}Supported \\ User Input\end{tabular}} & \multirow{-2}{*}{Method}                         & SAD                           & MSE                          & GRAD                          & CONN                          & SAD                           & MSE                          & GRAD                          & CONN                          \\ \hline
                                                                                  & ActiveMatting~\cite{active}                                     & 36.56                         & 0.93                         & 17.85                         & 37.46                         & 16.69                         & 0.49                         & 7.85                          & 16.73                         \\
                                                                                  & InteractiveMatting~\cite{click}                               & 30.47                         & 0.70                         & 15.82                         & 31.11                         & 14.06                         & 0.41                         & 7.02                          & 14.11                         \\
                                                                                  & DIIM~\cite{DIIM}                                             & 33.49                         & 0.81                         & 16.37                         & 32.04                         & 15.87                         & 0.43                         & 7.36                          & 15.89                         \\
                                                                                  & MatAny~\cite{matany}                                           & \multicolumn{1}{l}{30.49}     & 0.71                         & 15.89                         & 31.44                         & 14.01                         & 0.40                         & 7.00                          & 14.02                         \\
                                                                                  & \cellcolor[HTML]{EFEFEF}DFIMat(click trained)    & \cellcolor[HTML]{EFEFEF}28.63 & \cellcolor[HTML]{EFEFEF}0.67 & \cellcolor[HTML]{EFEFEF}15.2  & \cellcolor[HTML]{EFEFEF}28.49 & \cellcolor[HTML]{EFEFEF}13.74 & \cellcolor[HTML]{EFEFEF}0.39 & \cellcolor[HTML]{EFEFEF}6.49  & \cellcolor[HTML]{EFEFEF}13.77 \\
\multirow{-6}{*}{Click}                                                           & \cellcolor[HTML]{EFEFEF}DFIMat(mix trained)      & \cellcolor[HTML]{EFEFEF}28.01 & \cellcolor[HTML]{EFEFEF}0.66 & \cellcolor[HTML]{EFEFEF}16.84 & \cellcolor[HTML]{EFEFEF}27.36 & \cellcolor[HTML]{EFEFEF}13.59 & \cellcolor[HTML]{EFEFEF}0.38 & \cellcolor[HTML]{EFEFEF}6.47  & \cellcolor[HTML]{EFEFEF}13.62 \\ \hline
                                                                                  & SmartScribbles~\cite{smart}                                   & 26.59                         & 0.58                         & 16.01                         & 26.77                         & 12.42                         & 0.41                         & 6.27                          & 12.26                         \\
                                                                                  & FGI~\cite{flex}                                              & 27.34                         & 0.61                         & 16.85                         & 26.95                         & 13.63                         & 0.39                         & 6.51                          & 14.01                         \\
                                                                                  & UGDMatting~\cite{ugd}                                       & 30.95                         & 0.71                         & 17.90                         & 31.76                         & 14.58                         & 0.42                         & 7.39                          & 15.37                         \\
                                                                                  & MatAny~\cite{matany}                                           & \multicolumn{1}{l}{26.37}     & 0.58                         & 15.97                         & 26.82                         & 12.40                         & 0.41                         & 6.25                          & 12.51                         \\
                                                                                  & \cellcolor[HTML]{EFEFEF}DFIMat(scribble trained) & \cellcolor[HTML]{EFEFEF}24.14 & \cellcolor[HTML]{EFEFEF}0.50 & \cellcolor[HTML]{EFEFEF}15.82 & \cellcolor[HTML]{EFEFEF}24.19 & \cellcolor[HTML]{EFEFEF}12.22 & \cellcolor[HTML]{EFEFEF}0.38 & \cellcolor[HTML]{EFEFEF}6.25  & \cellcolor[HTML]{EFEFEF}12.23 \\
\multirow{-6}{*}{Scribble}                                                        & \cellcolor[HTML]{EFEFEF}DFIMat(mix trained)      & \cellcolor[HTML]{EFEFEF}23.86 & \cellcolor[HTML]{EFEFEF}0.49 & \cellcolor[HTML]{EFEFEF}15.79 & \cellcolor[HTML]{EFEFEF}23.80 & \cellcolor[HTML]{EFEFEF}12.07 & \cellcolor[HTML]{EFEFEF}0.37 & \cellcolor[HTML]{EFEFEF}6.18  & \cellcolor[HTML]{EFEFEF}11.99 \\ \hline
                                                                                  & MatAny~\cite{matany}                                           & \multicolumn{1}{l}{50.31}     & 2.36                         & 30.16                         & 50.52                         & 21.85                         & 0.93                         & 10.79                         & 22.44                         \\
                                                                                  & \cellcolor[HTML]{EFEFEF}DFIMat(box trained)      & \cellcolor[HTML]{EFEFEF}47.44 & \cellcolor[HTML]{EFEFEF}2.17 & \cellcolor[HTML]{EFEFEF}28.13 & \cellcolor[HTML]{EFEFEF}47.35 & \cellcolor[HTML]{EFEFEF}20.46 & \cellcolor[HTML]{EFEFEF}0.88 & \cellcolor[HTML]{EFEFEF}9.88  & \cellcolor[HTML]{EFEFEF}21.38 \\
\multirow{-3}{*}{Box}                                                             & \cellcolor[HTML]{EFEFEF}DFIMat(mix trained)      & \cellcolor[HTML]{EFEFEF}46.29 & \cellcolor[HTML]{EFEFEF}1.85 & \cellcolor[HTML]{EFEFEF}28.01 & \cellcolor[HTML]{EFEFEF}46.24 & \cellcolor[HTML]{EFEFEF}19.79 & \cellcolor[HTML]{EFEFEF}0.85 & \cellcolor[HTML]{EFEFEF}8.74  & \cellcolor[HTML]{EFEFEF}20.54 \\ \hline
                                                                                  & RIM~\cite{rim}                                              & 52.49                         & 2.94                         & 31.46                         & 52.87                         & 22.89                         & 0.97                         & 11.30                         & 23.33                         \\
                                                                                  & \cellcolor[HTML]{EFEFEF}DFIMat(text trained)     & \cellcolor[HTML]{EFEFEF}54.86 & \cellcolor[HTML]{EFEFEF}3.31 & \cellcolor[HTML]{EFEFEF}21.19 & \cellcolor[HTML]{EFEFEF}54.79 & \cellcolor[HTML]{EFEFEF}23.53 & \cellcolor[HTML]{EFEFEF}1.05 & \cellcolor[HTML]{EFEFEF}12.06 & \cellcolor[HTML]{EFEFEF}23.69 \\
\multirow{-3}{*}{Text}                                                            & \cellcolor[HTML]{EFEFEF}DFIMat(mix trained)      & \cellcolor[HTML]{EFEFEF}50.32 & \cellcolor[HTML]{EFEFEF}2.83 & \cellcolor[HTML]{EFEFEF}30.70 & \cellcolor[HTML]{EFEFEF}50.24 & \cellcolor[HTML]{EFEFEF}22.45 & \cellcolor[HTML]{EFEFEF}0.92 & \cellcolor[HTML]{EFEFEF}11.19 & \cellcolor[HTML]{EFEFEF}22.66 \\ \hline
Mix                                                                               & \cellcolor[HTML]{EFEFEF}DFIMat(mix trained)      & \cellcolor[HTML]{EFEFEF}22.89 & \cellcolor[HTML]{EFEFEF}0.47 & \cellcolor[HTML]{EFEFEF}15.54 & \cellcolor[HTML]{EFEFEF}22.73 & \cellcolor[HTML]{EFEFEF}11.77 & \cellcolor[HTML]{EFEFEF}0.36 & \cellcolor[HTML]{EFEFEF}5.98  & \cellcolor[HTML]{EFEFEF}11.80 \\ \hline
\end{tabular}}
\label{tab:Comparison}
\end{table}

\begin{figure}[t]
    \includegraphics[width=\linewidth]{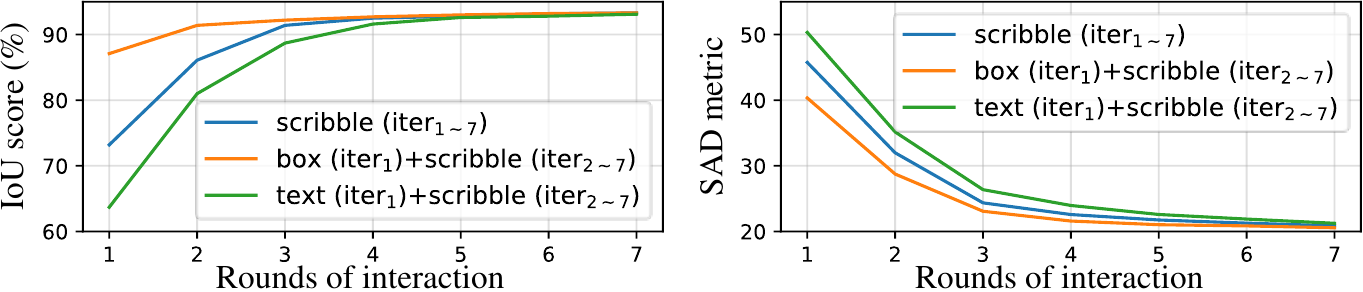}
    \captionsetup{font=footnotesize}
    \caption{Effect of different user inputs on the SMPMat dataset.}
     \label{fig:combine}
\end{figure}

\begin{figure*}[t]
    \centering
    \includegraphics[width=\linewidth]{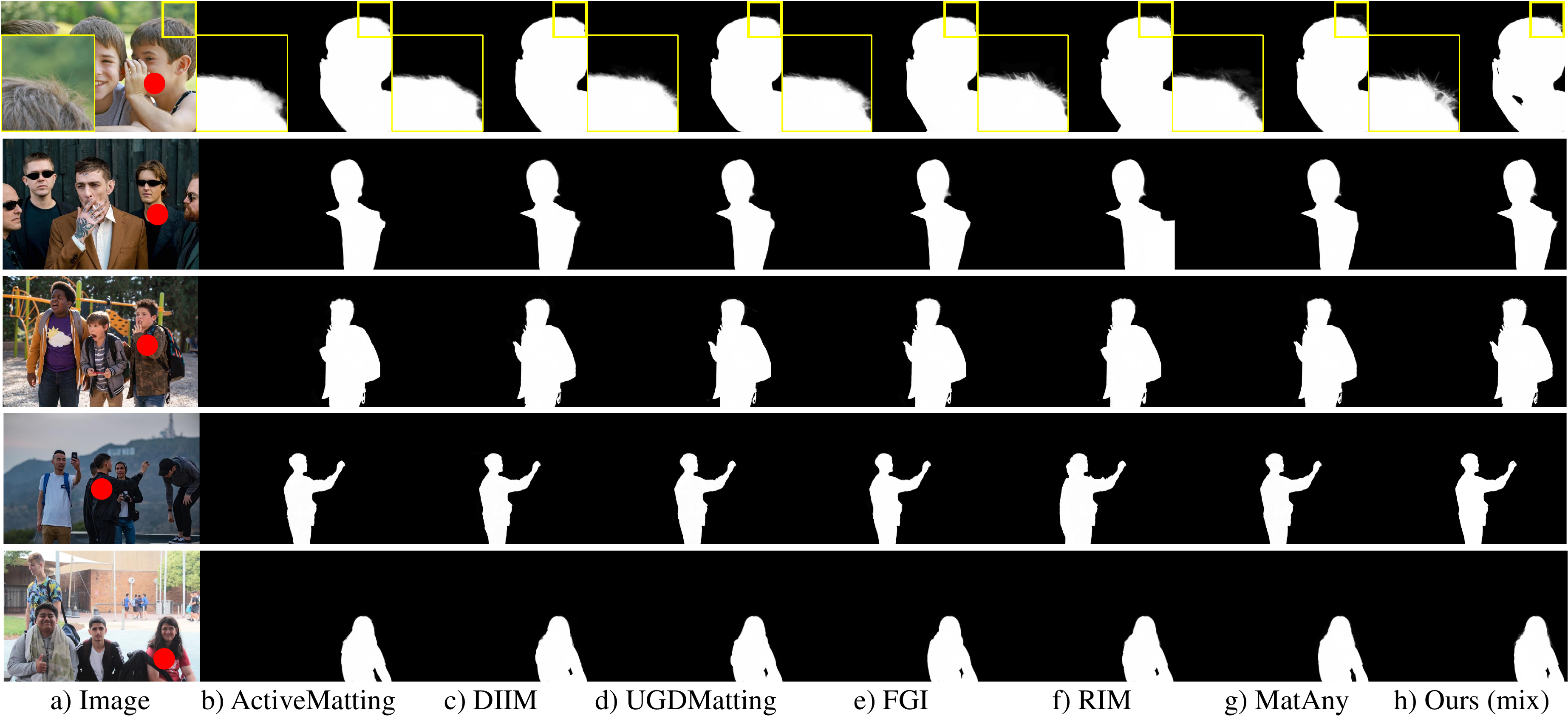}
    \caption{Qualitative comparisons among different methods. The red points in (a) indicate the target instances.}
    \label{fig:ex}
\end{figure*}

\noindent\textbf{Multi-type user input.} From the test results in \cref{tab:Comparison}, the following conclusions can be drawn: (1) Unlike previous methods that only support one type of input, our model enables mixed types of inputs. Experiments show that when trained under mixed user inputs, the performance of our model can be further enhanced, even with the same single-type input inference. This verifies the benefit of multi-type user input for model training, as they can give more complementary information. (2) As in the bottom line of \cref{tab:Comparison}, it can be seen that when applying mixed type inputs during inference, the performance of our model can be further improved. This further verifies the benefit of multi-type user inference, and it also makes user interaction more flexible.



\noindent\textbf{Analysis on user input choice.} Here we investigate the roles of different input types, aiming to provide valuable principles for users on more effective interaction. We assume only one input per interaction. We use IoU to measure the coarse-grained instance capture accuracy, and SAD to measure the fine-grained matting accuracy. As in Fig.~\ref{fig:combine}, box input can give a good start as its effectiveness for instance localization, scribble (we observe a similar role but slightly lower performance on click) is useful to refine local details, text is usually not-efficient. As a result, box at round 1 and scribble at the remaining iteration is an optimal choice for efficient user interaction.  




\noindent\textbf{Qualitative comparison.}
 We follow the same protocol in~\cref{sec: protocol} to conduct the experiment, as in~\cref{fig:ex}. The red points in (a) indicate the target instances. For our DFIMat, we only give the result from our full model (i.e., mix trained \& inference) in (h) due to the space limitation. More comparisons with other variants of DFIMat refer to our supplementary material. From~\cref{fig:ex}, it can be seen that DFIMat can accurately localize the target instance. More importantly, it shows superiority at perceiving fine-grained details: (1) Hair regions across all examples. (2) Other hollow body areas like the fingers or arm in row 2-5.


\begin{table}[t]
  \begin{minipage}{0.55\textwidth}
    \centering
    \caption{Analysis of the decoupled design.}
    \resizebox{\linewidth}{!}{
\begin{tabular}{c|c|cl|cc}
\Xhline{1pt}
\multirow{2}{*}{Setting} & Complexity & \multicolumn{2}{c|}{Semantic Capture} & \multicolumn{2}{c}{Matting}      \\
                         & GFLOPs     & \multicolumn{2}{c|}{NoC @ 90\%}        & SAD            & MSE             \\ \hline
Coupled Network          & 0.2186     & \multicolumn{2}{c|}{7.43}             & 25.19          & 0.53          \\
Coupled Traning          & 0.223      & \multicolumn{2}{c|}{6.82}             & 23.62          & 0.48          \\
\rowcolor{Gray}
DFIMat                   & 0.223      & \multicolumn{2}{c|}{\textbf{6.51}}    & \textbf{22.89} & \textbf{0.47} \\ \Xhline{1pt}
\end{tabular}}
\label{tab:Decouple}
 \end{minipage}
  \hfill
  \begin{minipage}{0.44\textwidth}
    \centering
    \caption{Effectiveness of CRM.}
    \resizebox{\linewidth}{!}{
\begin{tabular}{c|cl|cc}
\Xhline{1pt}
\multirow{2}{*}{Setting} & \multicolumn{2}{c|}{Semantic Capture} & \multicolumn{2}{c}{Matting}    \\
                         & \multicolumn{2}{c|}{NoC @ 90\%}        & SAD            & MSE           \\ \hline
w/o CRM          & \multicolumn{2}{c|}{7.93}             & 25.03          & 0.52          \\
\rowcolor{Gray}
w/ CRM                   & \multicolumn{2}{c|}{\textbf{6.51}}    & \textbf{22.99} & \textbf{0.47} \\ \Xhline{1pt}
\end{tabular}}
\label{tab:CRM}
  \end{minipage}
\end{table}

\begin{table}[t]
  \begin{minipage}{0.39\textwidth}
    \centering
    \captionof{table}{Ablation study on \\the encoder setting of MRN.}
    \resizebox{0.52\linewidth}{!}{
\begin{tabular}{c|cc}
\Xhline{1pt}
Branch    & SAD            & MSE           \\ \hline
Global  & 24.79          & 0.51          \\
Local        & 23.35          & 0.48          \\
\rowcolor{Gray}
Hybrid       & \textbf{22.89} & \textbf{0.47} \\ \Xhline{1pt}
\end{tabular}}
\label{tab:Encoder}
 \end{minipage}
  \hfill
  \begin{minipage}{0.6\textwidth}
    \centering
    \captionof{table}{Matting performance comparison under datasets using different data generation schemes.}
    \resizebox{\linewidth}{!}{
\begin{tabular}{c|cc|cc|cc}
\Xhline{1pt}
\multirow{2}{*}{Method}     & \multicolumn{2}{c|}{MG~\cite{MG}} & \multicolumn{2}{c|}{MatteFormer~\cite{mfor}} & \multicolumn{2}{c}{MRN}        \\
                            & SAD            & MSE           & SAD             & MSE            & SAD            & MSE           \\ \hline
Sys method in~\cite{inst} & 17.23          & 0.51          & 19.74           & 0.85           & {15.82} & {0.46} \\
\rowcolor{Gray}
Ours                        & \textbf{12.48}          & \textbf{0.43}          & \textbf{13.59}           & \textbf{0.38 }          & \textbf{11.77} & \textbf{0.36} \\ \Xhline{1pt}
\end{tabular}}
\label{tab:SMPMat}
  \end{minipage}
\end{table}

\subsection{Quantitative analysis on data synthesis method} \label{sec: quantitative_data_gen}

Here we give some quantitative evaluation of our proposed data synthesis pipeline. We compare it with the existing methods~\cite{inst}. We separately apply it and our method to synthesize a same amount of data (i.e., 45k instances) from model training. Then, we apply 3 different methods (i.e., MG~\cite{MG}, MatteFormer~\cite{mfor}, and MRN) to train on the synthesized data, and evaluate their testing performance on the HIM2K natural dataset. The result is listed in~\cref{tab:SMPMat}, it can be observed that when trained on the data synthesized by our method, the performance of different models is significantly and consistently better than trained on the data generated by existing method~\cite{inst}, which further verifies the superiority of our data synthesis pipeline.




\subsection{Ablation studies}
Here component effects are studied on the SMPMat dataset.


\noindent\textbf{Effect of the decoupled design.}
Here we evaluate the following settings: \textbf{a) Coupled Network} (ISCN with matting head); \textbf{b) Coupled Training} (ISCN + MRN but trained jointly); \textbf{c) DFIMat} 
 (Decoupled in both network and training). Tab. \ref{tab:Decouple} shows that both decoupled network and decoupled training have obvious performance gains (in both semantic capture and matting that align with our insight). Besides, the extra complexity from our decoupled design is neglectable (2\% in GFLOPS), which validates the effectiveness of our design.  We think the reason is that it can make the 2 independent tasks more focused and easier to be optimized, thus leading to a clear performance gain.

\noindent\textbf{Effect of the contrastive reasoning module (CRM).} From~\cref{tab:CRM}, it can be seen that with CRM, both semantic capture and matting performance improve by a noticeable margin, which shows the advantages of our design.

\noindent\textbf{Design choices of the matting refinement network (MRN).}~\cref{tab:Encoder} shows that our dual-stream design can enhance the performance. 


%% file: 6_conclusion.tex
\section{Conclusion and Limitations}
In this paper, we propose DFIMat, a decoupled framework that enables flexible interactive matting in multi-person scenarios, which consists of two modules, the interactive semantic capture network and the matting refinement network. DFIMat enables flexible and multi-type user input by encoding different inputs into a unified visual-semantic space, resulting in a more effective and user-friendly matting experience. Concerning the multi-round interaction requirement for practical usage, we also design a contrastive reasoning module to enhance cross-round refinement. To address the limitation from the perspective of data, we introduce a new synthetic data generation pipeline that can generate much more realistic samples than previous arts. A new large-scale dataset SMPMat is subsequently established. Extensive experiments verify the significant superiority of DFIMat while also providing valuable principles for efficient interaction. Despite its effectiveness, our method produces less accurate results when only coarse user input (e.g., box, text) is provided, and similar phenomena can also be observed in other methods. Besides, our SMP-Mat dataset does not include crowd scenes due to unrealistic generation results from base diffusion model~\cite{data1} in such cases. We will tackle those limitations in future works.
